\documentclass[10pt,twocolumn,letterpaper]{article}

\usepackage[pagenumbers]{cvpr} 

\usepackage{graphicx}
\usepackage{amsmath}
\usepackage{amssymb}
\usepackage{booktabs}

\makeatletter
\@namedef{ver@everyshi.sty}{}
\makeatother
\usepackage{tikz}
\usepackage{subcaption}
\newcommand*\circled[1]{\tikz[baseline=(char.base)]{
            \node[shape=circle,draw,inner sep=2pt, scale=0.65] (char) {#1};}}

\usepackage{listings}
\usepackage{algorithm,algorithmic}

\usepackage{multirow}

\usepackage[pagebackref,breaklinks,colorlinks]{hyperref}
\usepackage[accsupp]{axessibility}  

\usepackage[capitalize]{cleveref}
\crefname{section}{Sec.}{Secs.}
\Crefname{section}{Section}{Sections}
\Crefname{table}{Table}{Tables}
\crefname{table}{Tab.}{Tabs.}

\def\cvprPaperID{58} 
\def\confName{CVPR}
\def\confYear{2023}

\begin{document}

\title{DeepGEMM: Accelerated Ultra Low-Precision Inference on CPU Architectures using Lookup Tables}

\author{
Darshan C. Ganji \quad Saad Ashfaq \quad Ehsan Saboori \quad Sudhakar Sah \quad Saptarshi Mitra\\
MohammadHossein AskariHemmat \quad  Alexander Hoffman \quad Ahmed Hassanien\\
Deeplite Inc., Canada\\
{\tt\small \{darshan.ganji, saad, ehsan, sudhakar, saptarshi,}\\
{\tt\small mohammad, alexander.hoffman, ahmed\}@deeplite.ai}
\and
Mathieu Léonardon\\
IMT Atlantique, Lab-STICC, UMR CNRS 6285, F-29238 Brest, France\\
{\tt\small mathieu.leonardon@imt-atlantique.fr} 
}
\maketitle

\begin{abstract}
A lot of recent progress has been made in ultra low-bit quantization, promising significant improvements in latency, memory footprint and energy consumption on edge devices. Quantization methods such as Learned Step Size Quantization can achieve model accuracy that is comparable to full-precision floating-point baselines even with sub-byte quantization.
However, it is extremely challenging to deploy these ultra low-bit quantized models on mainstream CPU devices because commodity SIMD (Single Instruction, Multiple Data) hardware typically supports no less than 8-bit precision. To overcome this limitation, we propose DeepGEMM, a lookup table based approach for
the execution of ultra low-precision convolutional neural networks on SIMD hardware. The proposed method precomputes all possible products of weights and activations, stores them in a lookup table, and efficiently accesses them at inference time to avoid costly multiply-accumulate operations. Our 2-bit implementation outperforms corresponding 8-bit integer kernels in the QNNPACK framework by up to 1.74$\times$ on x86 platforms.
\end{abstract}

\section{Introduction}
\label{sec:intro}
Deep learning methods have achieved state-of-the-art performance on several computer vision tasks, but deploying these algorithms on mainstream CPU platforms is challenging due to cost and latency constraints \cite{sankaran2021neutrino}. Ultra low-bit quantization \cite{esser2019lsq, bhalgat2020lsq+, choi2018pact, choi2018sawb, zhou2016dorefa, li2021mqbench} presents an attractive option for reducing neural network inference costs. A 2-bit quantized model offers a theoretical model compression rate of 16$\times$ relative to the 32-bit floating-point (FP32) baseline, but achieving low latency inference with ultra low-bit models on general purpose processors (GPPs) remains an active area of research \cite{Tulloch2017, Cowan2020, Han2020}.

Deep learning workloads on CPUs are typically accelerated by exploiting data-level parallelism through SIMD programming. 
However, ultra low-bit deep learning operators can not be efficiently executed on these devices because sub-8-bit instructions are not generally supported in the vectorized instruction sets of mainstream CPU architectures including SSE/AVX instructions on x86  and Neon instructions on Arm. Therefore, to enable ultra low-precision model deployment on these GPPs, it is imperative to develop techniques that leverage the available SIMD instructions for ultra low-precision computations. 

\begin{table}
    \centering
    \caption{Top-1 accuracies on ImageNet dataset with LSQ.}
    \label{accuracies}
    \begin{tabular}{lccc}
        \toprule
        \multirow{2}{*}{\textbf{Model}} & 
        \multicolumn{3}{c}{\textbf{Accuracy@Precision}} \\ \cline{2-4}
        & 32-bit & 8-bit & 2-bit \\
        \midrule
        ResNet18 & 70.5\% & 71.1\% & 67.9\%  \\ 
        ResNet34 & 74.1\% & 74.1\% & 72.4\%  \\ 
        ResNet50 & 76.9\% & 76.8\% & 74.6\% \\
        VGG16 & 73.4\% & 73.5\% & 71.4\% \\
        \bottomrule
    \end{tabular}
\end{table}

Ultra low-bit quantization methods are highly competitive with industry standard 8-bit (INT8) quantization techniques, especially on classification tasks. For example, as shown in \cref{accuracies}, at 2 bits of precision, the state-of-the-art uniform quantization method LSQ \cite{esser2019lsq} achieves a 2.3\% and 2\% accuracy drop on the ImageNet dataset when quantizing the networks ResNet50 and VGG16, respectively. In comparison, the same method at INT8 precision results in an accuracy loss of 0.1\% and improvement of 0.1\% over the 32-bit 
full-precision baseline, respectively. The improved compression
and latency of the 2-bit model makes the mild accuracy
degradation tolerable in many use cases.

Moreover, ultra low-bit quantization can also be employed using a mixed precision approach to prevent the potentially significant accuracy loss resulting from quantizing all layers in a network. Sensitive layers can be kept at higher precision (FP32, FP16, INT8) and less sensitive layers can be quantized down to ultra low-precisions. HAWQ-V3 \cite{yao2020hawq} presents such a solution that determines the bitwidth per layer by solving an integer linear programming problem to balance the accuracy-performance tradeoff.

To realize the goal of ultra low-precision model deployment on GPPs, this paper proposes DeepGEMM, a set of convolution kernels that utilize the SIMD instructions in modern processors to perform lookup operations into precalculated tables. This method replaces expensive multiply-accumulate (MAC) operations prevalent in deep learning models with faster precomputed data retrieval from a lookup table (LUT) provided that the table is small enough to fit within the processor cache or registers. LUTs can be used for operations that require dot products such as convolutional and fully-connected layers which are the primary building blocks of convolutional neural networks (CNNs). Utilizing LUTs for these bottleneck layers can significantly accelerate inference on mainstream CPU platforms where some (mixed precision) or all of the layers in the network are quantized to ultra low-bit, based on the accuracy requirements. This paper makes the following contributions:

\begin{itemize}
    \item We propose a novel approach for extremely low-bit computations on CPUs with SIMD support that utilizes lookup tables to replace MAC operations. The dot product of two ultra low-bit input vectors can be performed by retrieving precomputed products of the input operands from a table stored in processor registers or cache eliminating costly arithmetic and enabling faster data access.
    \item We present DeepGEMM, a set of flexible ultra low-precision convolution operators compatible with uniform and non-uniform quantization techniques, for the x86 platform. The LUT based operators leverage efficient packing schemes and vectorization to minimize latency in ultra low-bit model deployment.
    \item We provide detailed performance breakdown and profiling results at the kernel, operator and model level. DeepGEMM offers substantial improvements over highly optimized GEMM libraries and state-of-the-art methods for ultra low-bit inference on x86.
\end{itemize}

The rest of the paper is structured as follows: Section \ref{Background} provides some background on quantization and prior works focusing on quantized inference on CPU architectures. Section \ref{Methodology} introduces DeepGEMM with a brief overview of the LUT approach and its different versions. Section \ref{Implementation_Details} takes a technical deep dive into the DeepGEMM algorithm detailing different packing schemes and the vectorized implementation. Section \ref{Experimental_Results} presents extensive experimental results with comparisons against optimized baselines and ultra low-bit techniques. Section \ref{Future_Work} discusses future enhancements and work in progress. Finally, Section \ref{Conclusion} concludes the paper.


\begin{figure*}[htp]
    \centering
    \begin{subfigure}{0.47\textwidth}
        \centering
        \includegraphics[width=\linewidth]{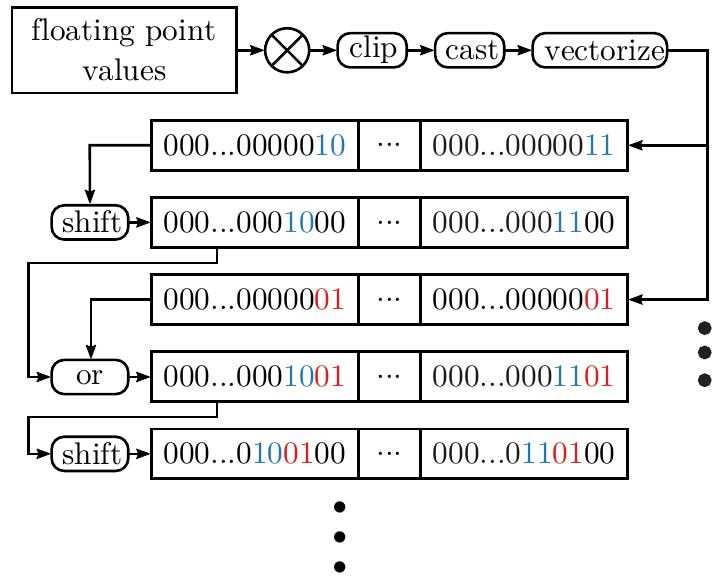}
        \caption{Vectorized packing}
        \label{packing}
    \end{subfigure}
    \hfill
    \begin{subfigure}{0.47\textwidth}
        \centering
        \includegraphics[width=\linewidth]{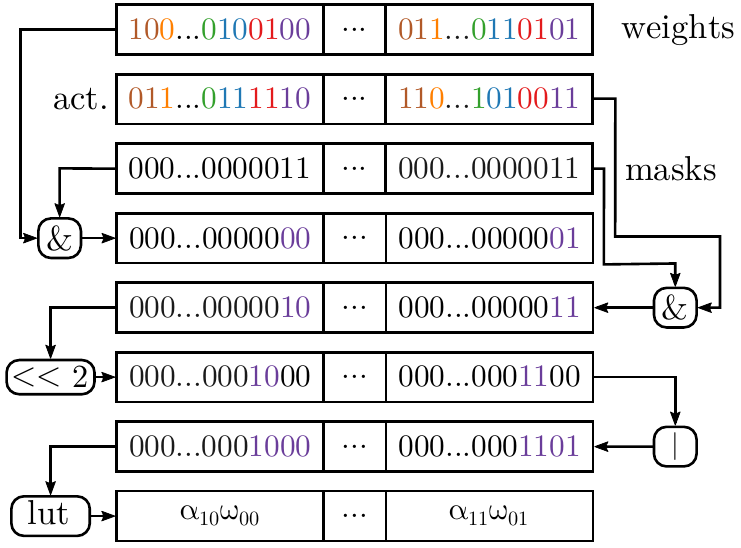}
        \caption{Vectorized unpacking}
        \label{unpacking}
    \end{subfigure}
    \caption{Weight and activation values in floating-point are quantized to ultra low-bit and cast to integer before being packed into higher percision data types in a vectorized manner using bitwise shift and OR operations in (a). Ultra low-bit weight and activation values are extracted from the packed data types and concatenated to form indices for LUT access using vectorized masking and bitwise operations in (b).}
\end{figure*}

\section{Background}
\label{Background}
\subsection{Quantization}
\label{sec:quantization}
In convolutional neural networks, the inputs to the convolutions are feature maps which are tensors of padded activations data that are convolved with weights. These activations and weights can be quantized. Quantization consists of mapping data from a larger set of potentially infinite real values to a smaller set of finite values represented using a countable number of bits. Typically, feature map and weight values are stored in floating-point format with 32 bits. It is possible to represent these values with fewer bits through quantization. The advantages of reducing the number of bits are  smaller memory footprint and lower computational complexity typically leading to increased performance with lower latency, higher throughput and improved energy efficiency. The term quantization (resp. dequantization) defines the process of converting 32-bit floating-point numbers into (resp. from) lower precision representations. There are two main families of quantization techniques: uniform and non-uniform.

In uniform quantization, the quantized values typically correspond to integers and there is a linear relation between intervals in the quantized and in the real domain. Two parameters are needed to define a particular quantization process, which are
a scaling factor $s \in \mathbb{R}$ and the zero-point $z \in \mathbb{R}$ which is the integer value to which the real value zero is mapped.
Let $x_q \in \{-2^{b-1}, -2^{b-1} + 1, ..., 2^{b-1}\}$ be the quantized version of $x \in \mathbb{R}$ that can be computed with the transformation given in \cref{quantization}.
\begin{equation}
\begin{split}
    x_q & = \text{quantize}(x) \\
     & = \text{clip}(\text{round}(s \cdot x + z, -2^{b-1}, 2^{b-1}-1))
    \label{quantization}
\end{split}
\end{equation}

The direct advantage of uniform quantization is that, operations on real values can be transformed into operations on integer values leading to optimized implementations on integer processing hardware. On the other hand, it is also possible to perform non-uniform quantization that generally offers higher accuracy by using floating-point values for the quantization levels. Although the transformation to integer arithmetic is 
lost, the mean quantization error is reduced by providing a better fit for the given distribution of weights and activations.

\subsection{Related Works}
\label{sec:related}
Prior efforts on low-bit implementation of CNNs on GPPs have primarily focused on 8-bit quantization. QNNPACK \cite{dukhan2018qnnpack} is a highly optimized library that is integrated within the PyTorch\cite{paszke2019pytorch} framework and provides high performance convolution kernels for 8-bit quantization targeting both x86 and Arm architectures. NCNN\cite{ncnn} and gemmlowp\cite{jacob2017gemmlowp} can also be used for efficient implementations of 8-bit quantized GEMM algorithms. Similarly, the CMSIS-NN framework\cite{Lai2018} includes optimized 8-bit convolution kernels for Arm Cortex-M processors.

Quantized inference on GPPs with fewer than 8-bits is more rare in the literature. \cite{Tulloch2017} proposes an implementation based on bitwise ANDs and popcount instruction to perform the elementary operations of the convolutions. Indeed, if weights and activations are represented with a single bit, a multiplication can be implemented as an AND, and accumulation can be performed using the popcount operation. By adding shifts to the equation, one can also support computations on 2-bit or higher precision data enabling efficient data parallelism and lower latency compared to standard higher precision operators.

This bit-serial approach with bitwise ANDs and popcount operations is also used in \cite{Cowan2020} and \cite{ashfaq2022lowbitruntime} where bit-serial kernel implementations for Arm-based targets are provided within the TVM \cite{chen2018tvm} machine learning compiler framework. This allows the algorithms to take advantage of existing compiler optimizations including loop tiling, loop unrolling, vectorization and multicore parallelization. The method is tested on a ResNet18 network, and the throughput gains are significant, especially for 1-bit quantization ($6\times$ over FP32).

Another framework that increases the performance of GEMM processing at ultra low-precision is ULPPACK~\cite{ulppack}. Depending on the number of bits, this implementation is shown to achieve better performance than the bit-serial method. The principle is to pack multiple sub-byte weight and activation values  within an 8-bit integer. Then a regular 8-bit multiply operation produces the expected dot-product result. This principle can be further extended to 16-bit and 32-bit multiply operations with even more inputs. Full networks are implemented using this technique, and it is shown that ULPPACK outperforms other low-bit implementations such as QNNPACK, gemmlowp, and the bit-serial version at certain bitwidths.

\cite{Han2020} have also proposed an implementation of extremely low-bit quantization of CNNs. The performance gains are achieved with a clever usage of MAC operations that are vectorized using the Arm Neon instruction set, reducing the number of cycles needed to perform a single convolution. Throughput gains are significant and are even observed for 4-8 bits, while the bit-serial approach in \cite{Cowan2020} had only reported gains with 1-3 bits.

Bitflow \cite{hu2018bitflow} implements binary network on CPUs. It describes a code generator that produces 1-bit quantized operators and uses data-level and thread-level parallelism to achieve good performance. A PressedConv layer is introduced that performs bit-packing along the channel dimension, and then applies bitwise XORs and popcount operations. A feature that is specific to this implementation is the use of AVX512 instructions on Xeon Phi architecture.

\section{Methodology}
\label{Methodology}
\subsection{Overview}
\begin{figure}
    \centering
    \includegraphics[width=\linewidth]{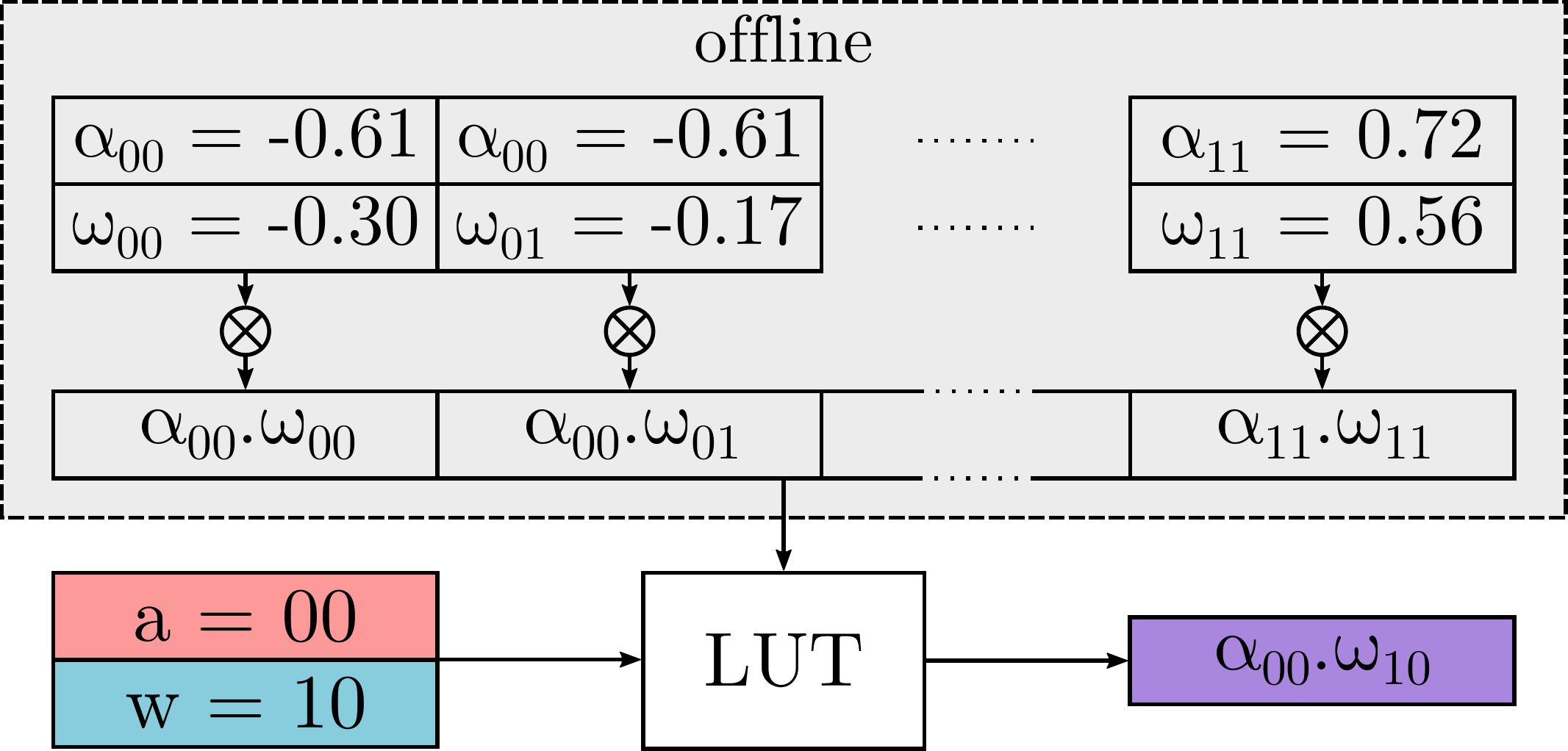}
    \caption{Lookup table accessed to retrieve precomputed products of ultra low-bit activation and weight values that are represented by the corresponding indices.}
    \label{fig:fp lookup}
\end{figure}

The overall DeepGEMM algorithm is composed of three primary steps: packing, unpacking and lookup table access. Packing involves loading multiple low-bit values within a standard higher precision data type. \cref{packing} depicts the vectorized packing of 2-bit input values within larger data types. Unpacking extracts the low-bit values from the packed representation as illustrated in \cref{unpacking}. Both packing and unpacking steps only require inexpensive bitwise masking and shifting operations. Unpacked low-bit weight and activation values are concatenated to construct indices that are used for LUT access as shown in \cref{fig:fp lookup}.

\subsection{LUT Versions}
We developed two different implementations of our LUT approach: the first uses SIMD operations on a lookup table with $16$ entries (LUT-16) while the second uses a lookup table with $2^{16}$ entries (LUT-65k). In both cases, the weight and activation values are quantized to 2 bits. The implementations target x86 CPUs with AVX2 support using 256-bit vector registers. For both versions, the lookup table entries are 8-bit values implicitly assuming that the result of the MAC operations will not overflow 8 bits. If required, higher precision data types can be chosen for the lookup table entries to account for larger accumulation results to be stored in the table.

\begin{figure}
    \centering
    \includegraphics[width=\linewidth]{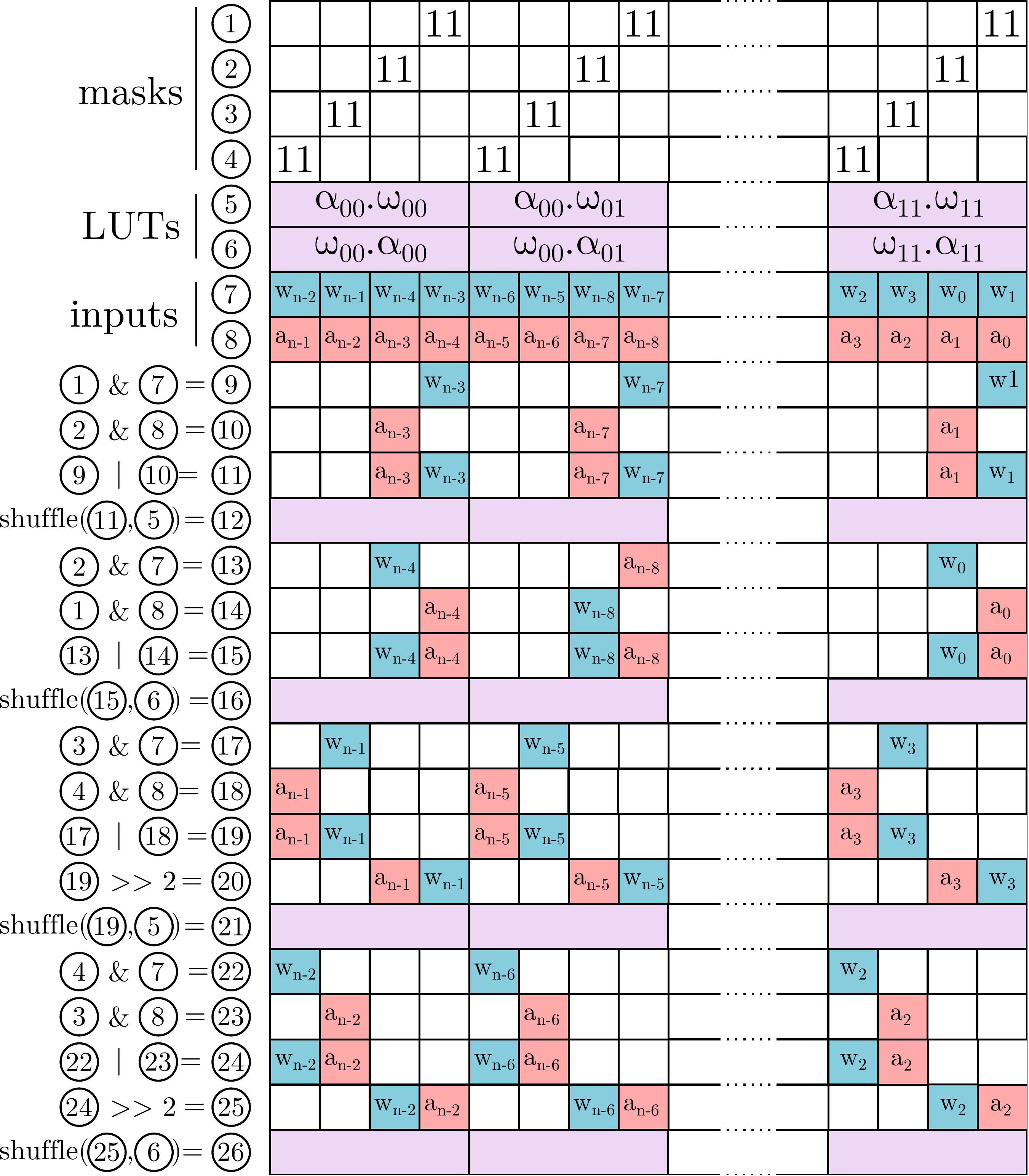}
    \caption{Vectorized lookup with LUT-16.}
    \label{fig:lut16 lookup}
\end{figure}

\subsection*{LUT-16}
In the LUT-16 version shown in \cref{fig:lut16 lookup}, two 128-bit lookup tables are stored in a single 256-bit vector register. The values stored in the LUT are all possible results of $w \cdot a$, with $w \in \{\omega_{00}, \omega_{01},\omega_{10}, \omega_{11}\}$ and $a \in \{\alpha_{00}, \alpha_{01},\alpha_{10}, \alpha_{11}\}$. A notable trick is that the weights in \circled{7} are reordered so that after masking, the resulting vectors \circled{9} and \circled{10} can be directly combined with an OR operation, without additional shifting needed. And this is cost-less at inference time, because the rearrangement of weights can be performed offline. The resulting values in \circled{11}, \circled{15}, \circled{20} and \circled{25} represent the indices that are used to retrieve the precomputed products from the LUTs stored in \circled{5} and \circled{6} with the AVX2 shuffle instruction for table lookup.

This process produces 4, 8-bit dot products in vector registers \circled{12}, \circled{16}, \circled{21} and \circled{26}. A vectorized sum of the 8-bit elements is performed across these registers. The final operation is a horizontal addition of these values.
The AVX2 implementation of this reduction is detailed in \cref{lst:label1}.

\begin{lstlisting}[caption=Reduction after interleaving and lookup,label={lst:label1}, basicstyle=\footnotesize, breaklines=false]
__m128i a = _mm256_castsi256_si128(sum);
__m128i b = _mm256_extracti128_si256(sum, 1);
__m128i d = _mm_add_epi64(b, a);
__m128i e = _mm_shuffle_epi32(d, 238);
__m128i f = _mm_add_epi64( e, d);
out = _mm_cvtsi128_si64(f);
\end{lstlisting}

Upon close inspection, it can be inferred that the number of instructions to perform the convolution with the LUT approach is actually comparable to the FP32 baseline. For each pair of weight and activation values, we perform an operation in both cases: a shuffle for our ultra low-bit LUT method and a multiplication for the standard FP32 method. When loading vector registers with weight and activation inputs, the number of values loaded in a register is given by the vector register size $(R)$ divided by the data size. For 2-bit quantized values with the LUT approach, $R/2$ values can be loaded in a register whereas in the case of FP32 data, only $R/32$ values can be loaded ($R/8$ in the case of 8-bit quantized data). Therefore, in addition to the latency of shuffle operation being lower than multiplication, the observed gains with the LUT method also result from substantially fewer exchanges between the cache and registers.

\subsection*{LUT-65k}
In the LUT-65k version, we create a lookup table with $2^{16}$, 8-bit elements. The elements correspond to all possible combinations of the dot product between 4, 2-bit weight and 4, 2-bit activation values. The index into this table is 16-bit wide and is constructed by concatenating the 8 bits from the 4 weights with the 8 bits from the 4 activations. This greatly simplifies the unpacking step as the 8 bits of weights and activations can be easily interleaved using existing 8-bit vectorized instructions removing the need for explicit masking and shifting operations to extract the relevant values. With a total size of 64 KB, the lookup table can be stored in cache, as it easily fits within a typical L2 cache on modern processors.

\begin{table}
\centering
\caption{Scaling LUT-16 to larger bitwidths.}
\label{Scaling LUT to larger bitwidths}
\begin{tabular}{llll}
\toprule
\textbf{Bitwidth} & 2 & 3  & 4 \\
\midrule
\textbf{Index bitwidth} & $2+2 = 4$ & $3+3 = 6$ & $4+4 = 8$ \\
\midrule
\textbf{LUT entries} & $2^4 = 16$ & $2^6 = 64$ & $2^8 = 256$ \\
\midrule
\textbf{LUT size} & 128 bits & 512 bits & 2048 bits \\ 
\midrule
\textbf{AVX2 registers} & 1 & 2 & 8 \\ 
\midrule
\textbf{Fits in L1 cache} & $\checkmark$ & $\checkmark$ & $\checkmark$ \\
\bottomrule
\end{tabular}
\end{table}

\subsection{Scalability to Larger Bitwidths}
Currently DeepGEMM only supports 2-bit quantized models, but the LUT-16 version can be easily extended to larger bitwidths with the set of changes listed in \cref{Scaling LUT to larger bitwidths}. For 3-bit, we require a 6-bit index into the LUT resulting in 64 entries. Assuming each entry is an 8-bit integer, this would require 64 bytes or 512 bits so the LUT can be stored in 2, 256-bit AVX2 registers or the L1 cache. The number of packing and unpacking instructions remain the same as the 2-bit version but the LUT access time will slightly increase due to accessing entries within a larger table. Similarly for 4-bit, an 8-bit index is needed to access a 256 entry LUT that requires 256 bytes or 2048 bits for storage. The LUT can be stored in 8 AVX2 registers or in the L1 cache. Instructions for packing are the same as the 2-bit version but less instructions are required for unpacking since the 8-bit index can be supported by the standard INT8 data type eliminating some masking and shifting operations. However, the overhead for accessing the LUT will be slightly higher due to retrieving results from a larger table spread across multiple vector registers.

\section{Implementation Details}
\label{Implementation_Details}
\subsection{Packing Schemes}
\begin{figure}
    \centering
    \includegraphics[width=0.47\textwidth]{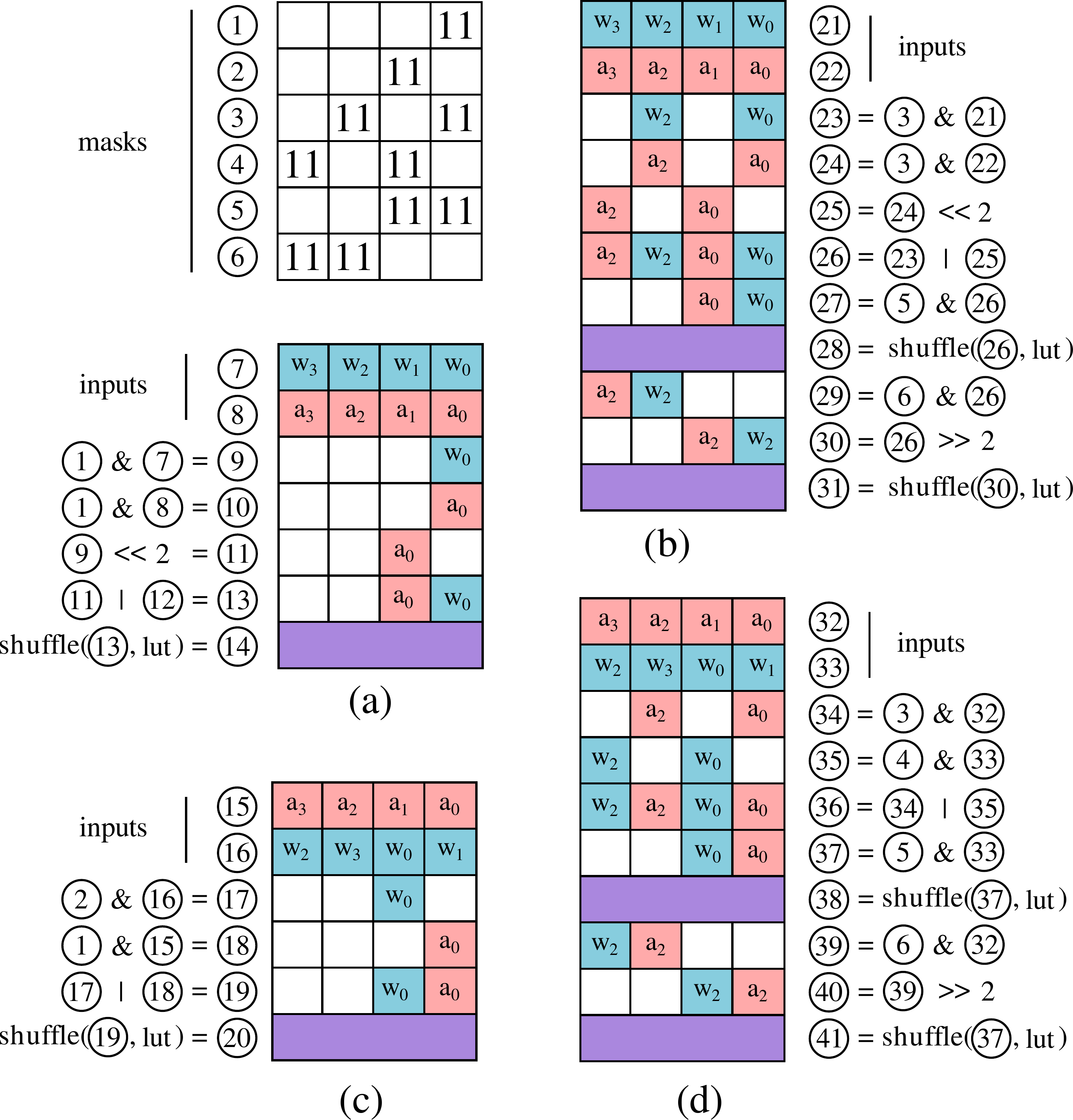}
    \caption{Packing schemes and corresponding operations for unpacking.}
    \label{fig:packing schemes}
\end{figure}

We experimented with several packing schemes in order to optimize the unpacking step. \cref{fig:packing schemes} displays 2 different packing schemes and 2 unpacking processes for each scheme. The first packing scheme used in (a) and (b) is the most naive one. The inputs \circled{7} and \circled{8} are naturally ordered. In order to unpack them, 4 operations are needed to put 2 of the inputs in the correct order, before being used as index to the lookup operation, implemented with a shuffle. To generate a second output, the same process would be needed but with an additional shift before doing the final lookup step.  In figure (b), a small improvement is proposed with the same packing scheme, but using different masks (\circled{3}, \circled{5} and \circled{6}), in order to reduce the number of operations needed to unpack two pairs of weights and activations. Further improvement can be performed with the packing scheme described in schemes (c) and (d) (see \circled{16}). With this new arrangement of weights performed offline, there is one less shift operation to be performed in scheme (c) compared to scheme (a) to produce 1 output. Finally, scheme (d) uses both improvements, generating two pairs of weights and activation as in (b), with the packing scheme used in (c) to reduce the number of shifts needed. \cref{Number of instructions for different packing schemes} lists the average number of instructions required to retrieve a single entry from the LUT for one weight-activation pair with these different schemes.

\begin{table}
    \centering
    \caption{Average number of instructions needed per output for different packing schemes.}
    \label{Number of instructions for different packing schemes}
    \begin{tabular}{lcccc}
        \toprule
        \textbf{Instruction} & \textbf{a}  & \textbf{b} & \textbf{c} & \textbf{d} \\
        \midrule
        AND & 2 & 2  & 2  & 2 \\ 
        Shift & 1.5 & 1 & 0.5  & 0.5\\ 
        OR & 1  & 0.5 & 1  & 0.5\\ 
        Shuffle & 1 & 1 & 1  & 1\\
        \midrule
        Total & 5.5 & 4.5 & 4.5 & 4\\
        \bottomrule
    \end{tabular}
\end{table}

\subsection{Vectorized Algorithm}
We are able to execute 32 lookups at a time within a 256-bit vector register using the AVX2 shuffle instruction. \cref{algo:deepgemm} shows simplified pseudocode for the vectorized DeepGEMM algorithm. The precomputed products are stored in the LUT at line 4. The prepacked activation and weight values are loaded into 256-bit  vector registers in lines 5 and 6, respectively. The unpacking of inputs to generate the index for LUT access happens at line 9. Finally, the lookup table is accessed to retrieve the precomputed product of the input values in line 10. The sum reduction of the LUT values is the last step before storing the accumulated dot product results. Unlike Neon, AVX2 does not offer an instruction that can be utilized for the horizontal vector sum; the group of instructions that were utilized for this reduction step are given in \cref{lst:label1}.

\begin{algorithm}
\caption{DeepGEMM algorithm as pseudocode}\label{algo:deepgemm}
\begin{algorithmic}[1]
\STATE $mask$: Register that extracts the 2 LSB bits from 8 bit elements of a vector register.
\STATE $lookup\_table$: Lookup table containing every combination of multiplication of two 2-bit values.
\STATE $addr[V]$: Returns the address of variable $V$.

\STATE $lut \gets addr[lookup\_table]$
\STATE $vec_a \gets addr[act\_tensor]$
\STATE $vec_w \gets addr[w\_tensor]$

\FOR{$i \gets 0$ to $4$}
    \STATE $shift = i*2$
    \STATE $indx= ((vec_w>>shift)\& mask)|((vec_a>>shift)\&mask)$
    \STATE $res = shuffle(lut, indx)$
\ENDFOR
\STATE $reduction\_sum$ 
\end{algorithmic}
\end{algorithm}

  
                                    




\begin{figure}
    \centering
    \begin{subfigure}{0.47\textwidth}
        \centering
        \includegraphics[width=\linewidth]{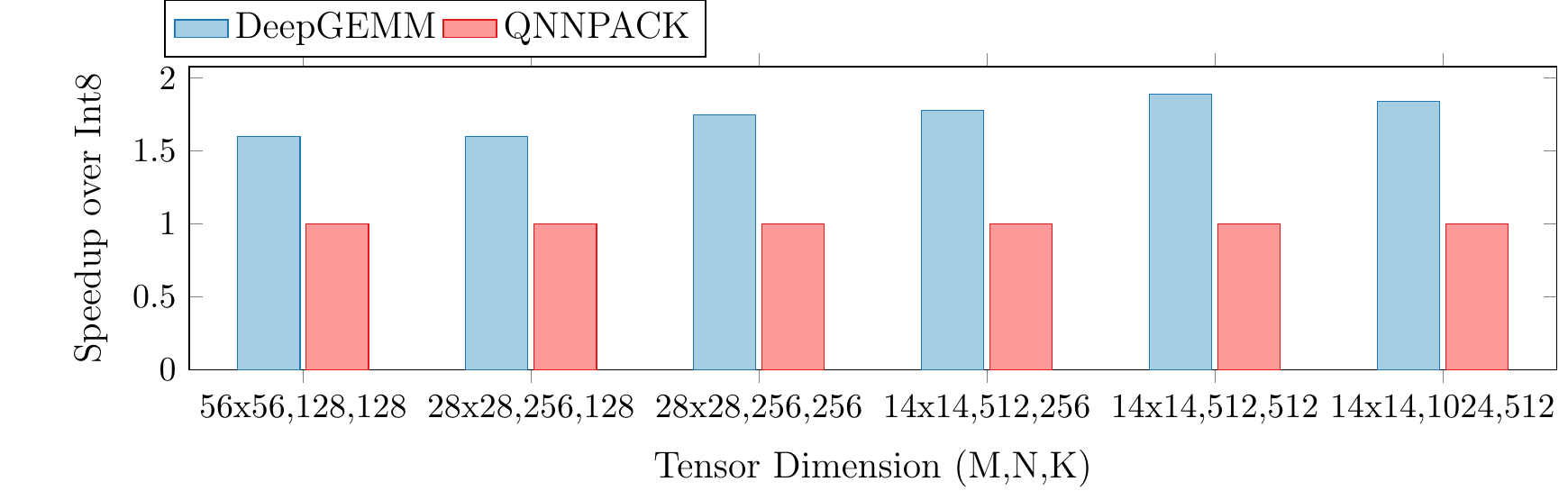}
        \caption{MobileNetV1}
        \label{fig:mobilenetv1}
    \end{subfigure}
    \begin{subfigure}{0.47\textwidth}
        \centering
        \includegraphics[width=\linewidth]{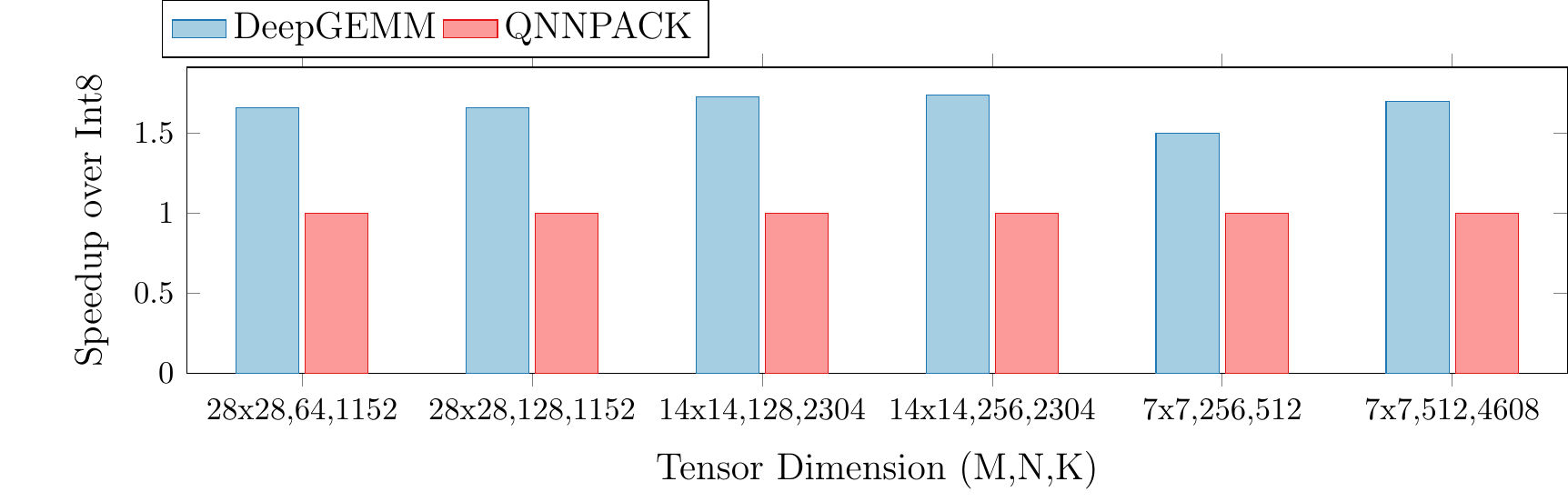}
        \caption{ResNet18}
        \label{fig:resnet18}
    \end{subfigure}
    \begin{subfigure}{0.47\textwidth}
        \centering
        \includegraphics[width=\linewidth]{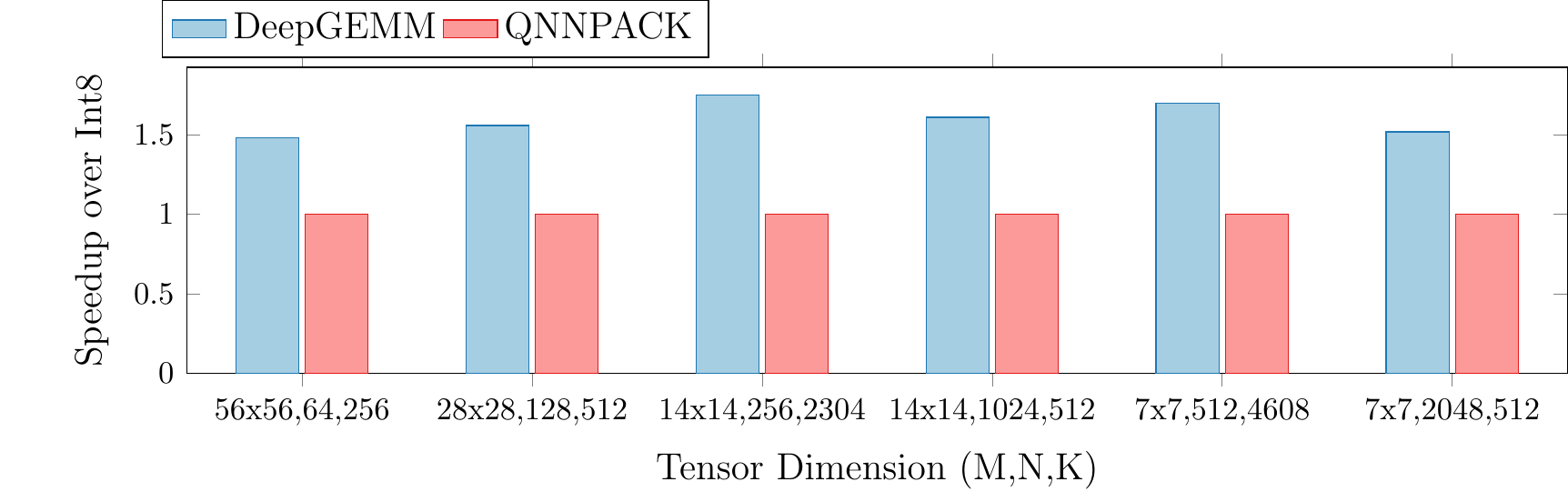}
        \caption{ResNet50}
        \label{fig:resnet50}
    \end{subfigure}
    \caption{Per-layer speedups over QNNPACK INT8.}
    \label{fig:per layer speedups}
\end{figure}

\section{Experimental Results}
\label{Experimental_Results}
\subsection{Operator Profiling}
On the x86 platform (Intel i7 9700k@3.6GHz), we compare the performance of DeepGEMM on various classification models against optimized INT8 kernels in the QNNPACK library. The results on a subset of the convolution layers are given in \cref{fig:per layer speedups} with each subfigure corresponding to a classification model. The horizontal axis gives the dimensions of each layer in $(M, N, K)$ format representing an $(M, N) \times (N, K)$ GEMM computation. The per-layer speedup increases with higher values of $K$ as the DeepGEMM kernel is vectorized along the $K$ dimension. The overall geomean speedups across all the layers tested are presented in \cref{per layer geomean speedups}. DeepGEMM achieves significantly lower latencies compared to QNNPACK with geomean speedups of 1.74$\times$, 1.64$\times$, 1.67$\times$ and 1.57$\times$ on MobileNetV1, ResNet18, ResNet34 and ResNet50 layers, respectively.

\begin{table}
    \centering
    \caption{Geomean speedups across convolution layers over QNNPACK INT8.}
    \label{per layer geomean speedups}
    \begin{tabular}{lc}
        \toprule
        \textbf{Model} & \textbf{Geomean Speedup} \\
        \midrule
        MobileNetV1 & 1.74$\times$ \\ 
        ResNet18 & 1.64$\times$ \\ 
        ResNet34 & 1.67$\times$ \\ 
        ResNet50 & 1.57$\times$ \\
        \midrule
        Average & 1.66$\times$ \\
        \bottomrule
    \end{tabular}
\end{table}

\subsection{End-to-end Profiling}
\begin{figure*}
    \centering
    \includegraphics[width=0.80\linewidth]{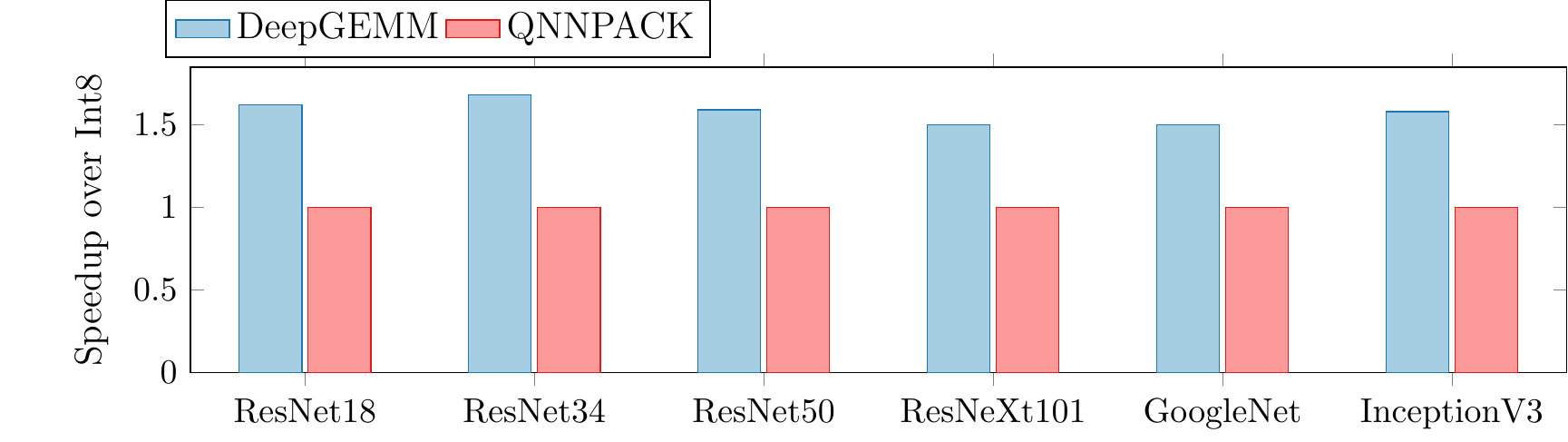}
    \caption{End-to-end speedups over QNNPACK INT8.}
    \label{fig:end-to-end speedups}
\end{figure*}

End-to-end inference results for six convolutional neural networks tested with QNNPACK and DeepGEMM kernels are presented in \cref{fig:end-to-end speedups} and \cref{end to end performance}. All convolution layers in the networks were quantized to either 8-bit for QNNPACK or 2-bit for DeepGEMM. The results include the execution time for packing, quantization and dequantization of activations while the packing and quantization of weights was handled offline since the parameter values are available after training. The results demonstrate significant end-to-end speedups of up to 1.68$\times$ with DeepGEMM over QNNPACK, specially on models such as ResNet, where the convolution layers form the primary bottleneck during model execution. 

The latency improvements enabled with DeepGEMM are complemented by advances in ultra low-bit quantization methods where minimal accuracy degradation is observed across entire networks with all layers converted to ultra low-precision. This is demonstrated in \cref{accuracies} that gives the Top-1 accuracies for the ResNet-based models at 32, 8 and 2 bits with the quantized models trained using LSQ \cite{esser2019lsq}. The 2-bit ResNet18, ResNet34 and ResNet50 models only incur accuracy drops of 2.6\%, 1.7\% and 2.3\% compared to the full-precision baselines, respectively. The accuracy losses relative to the 8-bit baselines are 3.2\%, 1.7\% and 2.2\%, respectively. The higher inference performance realized with DeepGEMM coupled with the considerable memory savings achieved through ultra low-bit model representations can offset the minor accuracy dips in several applications.   

\subsection{Comparison with Ultra Low-bit Methods}
State-of-the-art methods for ultra low-bit inference on GPPs including  bit-serial \cite{Cowan2020} and ULPPACK \cite{ulppack} primarily focus on kernels for Arm CPU targets whereas DeepGEMM provides a set of kernels for the x86 platform with porting to Arm architecture still under development. ULPPACK does offer some x86 performance results but no low-level implementation details are provided so it was not possible to reproduce the reported results on our platform. A comparison may be performed by using the per-layer throughput numbers from ULPPACK on a subset of MobileNetV1 layers at 2-bit; ULPPACK achieves a geomean speedup of 1.77$\times$ whereas DeepGEMM achieves a geomean speedup of 1.74$\times$ over QNNPACK as shown in \cref{fig:mobilenetv1} and \cref{per layer geomean speedups}. In addition to offering competitive performance to SOTA, DeepGEMM also provides more flexibility for quantization and input representations.

The LUT can store either integer or floating-point values. Floating-point entries in the LUT that represent the dot products of input activation and weight values make DeepGEMM compatible with non-uniform quantization. Non-uniform quantizers typically use floating-point values for the quantization levels producing models with higher accuracy compared to uniform quantizers. LCQ \cite{yamamoto2021lcq} uses non-uniform quantization to outperform the SOTA LSQ method by up to 1.3\% in image classification, demonstrating the potential performance improvements enabled by DeepGEMM. Bit-serial and ULPPACK only work with integer values and are therefore not suitable for non-uniform quantization.

The LUT can store either unsigned or signed data. This allows the input activations and weights to be encoded as unipolar (unsigned) or bipolar (signed). ULPPACK only supports unsigned data and therefore requires additional operations to be added before and after the GEMM computation to accommodate signed inputs such as neural network weights. Although the bit-serial approach natively supports both unipolar and bipolar data representations without the addition of any operators to the network, the bipolar case requires extra popcount instructions in the dot product calculations \cite{Cowan2020}. DeepGEMM offers the distinct advantage of providing identical latency regardless of the sign of the input data enabling the same level of performance for higher accuracy bipolar quantization techniques \cite{esser2019lsq, choi2019pactsawb, cai2017hwgq} relative to purely unipolar methods.

%
%

Finally, since the LUT stores precomputed results, and quantization constants including scaling factors are known at compile-time, this potentially enables the fusion of quantize, convolution and dequantize operators by storing the combined results of the sequence of these layers in the LUT. This can remove the overhead of activation quantization and dequantization which can be significant for some layers.

\begin{table}
    \centering
    \caption{End-to-end speedups over QNNPACK INT8.}
    \label{end to end performance}
    \begin{tabular}{lc} 
        \toprule
        \textbf{Model} & \textbf{End-to-end Speedup} \\
        \midrule
        ResNet18 & 1.62$\times$ \\ 
        ResNet34 & 1.68$\times$ \\ 
        ResNet50 & 1.59$\times$ \\
        ResNeXt101 & 1.50$\times$ \\
        GoogleNet & 1.50$\times$ \\
        InceptionV3 & 1.58$\times$ \\
        \midrule
        Average & 1.58$\times$ \\
        \bottomrule
    \end{tabular}
\end{table}

\begin{figure}
    \centering
    \includegraphics[width=0.90\linewidth]{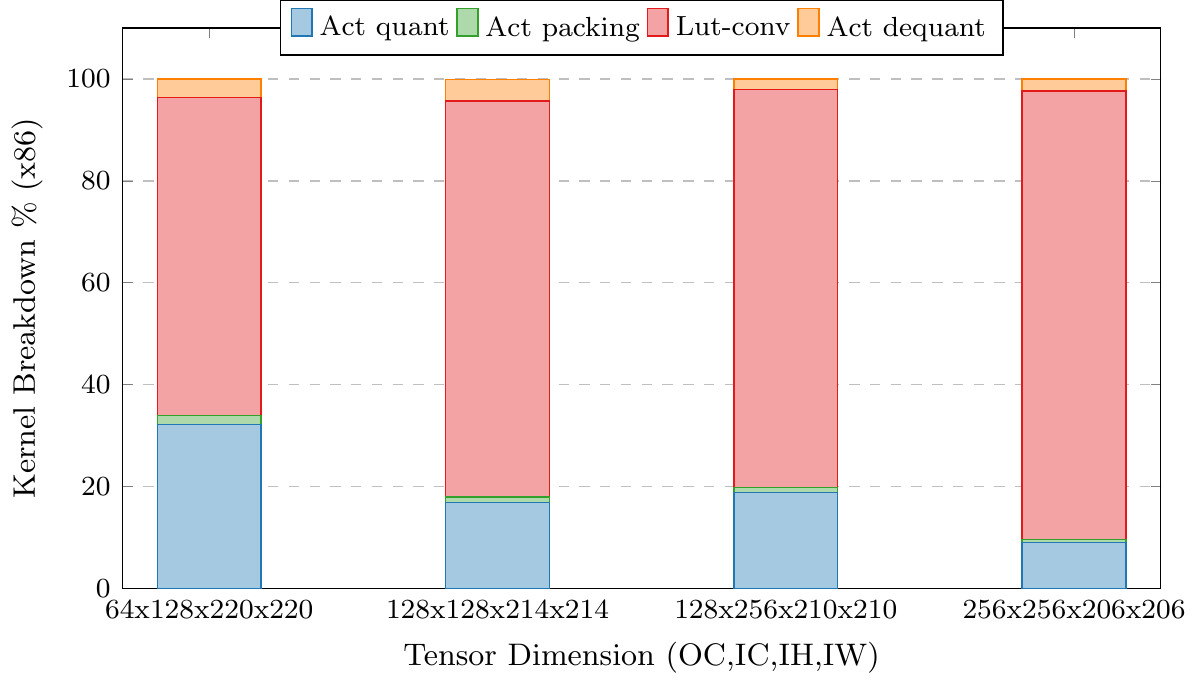}
    \caption{Kernel profiling on x86 platform.}
    \label{fig:x86 profiling}
\end{figure}

\begin{figure}
    \centering
    \includegraphics[width=0.90\linewidth]{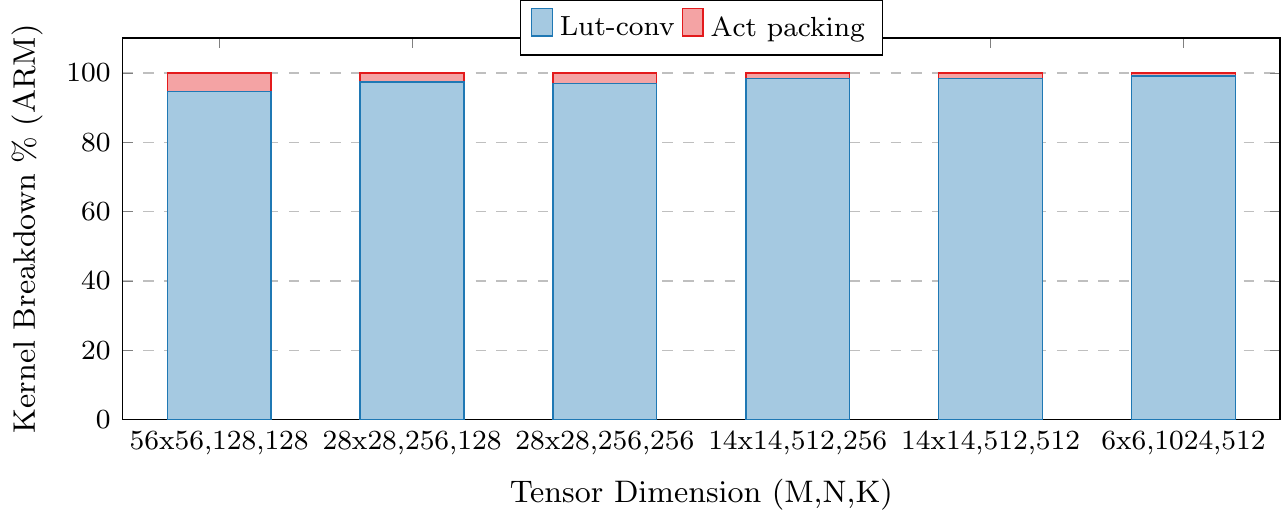}
    \caption{Kernel profiling on Arm platform.}
    \label{fig:arm profiling}
\end{figure}

\subsection{Low-Level Kernel Profiling}
\cref{fig:x86 profiling} provides a detailed breakdown of the execution times for individual layers with DeepGEMM on the x86 platform. Operations for a single convolution layer can be categorized into activation quantization, activation packing, convolution with LUT (includes unpacking, lookup and accumulation) and activation dequantization. We used the ONNX Runtime profiling tool to get the latencies for these four stages and also used the Intel VTune\textsuperscript{\texttrademark} Profiler to compare the latencies of the three steps within the LUT convolution stage. Activation packing and dequantization take a minor amount of the overall execution time. For small layers, activation quantization can add considerable overhead, but this can be addressed by the fusion of operators as discussed in the last point when comparing to SOTA methods above. In all cases, it can be observed that the Lut-Conv step that includes unpacking, lookup and accumulation, is the primary bottleneck so DeepGEMM should still be performant when the overhead of activation packing, quantization and dequantization is considered. The VTune\textsuperscript{\texttrademark} Profiler results further reveal that the unpacking step within Lut-Conv
consumes about 80\% of the overall execution time. Since unpacking has to be performed repeatedly on the packed data and requires multiple bitwise instructions, it presents a significant overhead compared to the lookup operation. Interestingly, even though table lookup is the defining feature of DeepGEMM, it takes substantially less time compared to both the input packing and unpacking steps promising considerable performance improvements in the future with more efficient unpacking schemes.

\section{Future Work}
\label{Future_Work}
As stated previously, the unpacking stage is currently the primary bottleneck in the DeepGEMM algorithm. We are exploring more efficient packing methods to reduce the number of instructions required for unpacking. We are also experimenting with operator fusion before and after convolution to store precomputed results for a sequence of operators in the LUT to remove the overhead resulting from activation quantization. In addition to our x86 LUT kernels, we also worked on extending DeepGEMM to Arm and tested it on the Raspberry Pi 4B device (4$\times$Arm Cortex-A72@1.5GHz) as shown in \cref{fig:arm profiling}. However, Neon lacks a 128-bit vectorized instruction for table lookup similar to the AVX2 shuffle instruction so our current Arm implementation does not offer competitive performance. 

\section{Conclusion}
\label{Conclusion}
In this work, we introduce DeepGEMM, a lookup table based approach for CPUs that replaces the costly multiply-accumulate arithmetic in dot product calculations with simpler indexing operations into preconstructed tables for computing ultra low-precision layers in convolutional neural networks. The faster latency of table lookup coupled with substantially fewer memory accesses for packed ultra low-bit data results in significant performance gains over optimized baselines. Compared to prior works on sub-byte operators, DeepGEMM offers greater flexibility with support for uniform and non-uniform quantization methods where unsigned and signed, integer or floating-point results can be precomputed and stored in the lookup table. We implement vectorized DeepGEMM kernels for x86 platforms that outperform optimized 8-bit baselines by up to 1.74$\times$.

{\small
\bibliographystyle{ieee_fullname}
\bibliography{main}

\begin{thebibliography}{10}\itemsep=-1pt

\bibitem{ashfaq2022lowbitruntime}
Saad Ashfaq, MohammadHossein AskariHemmat, Sudhakar Sah, Ehsan Saboori, Olivier
  Mastropietro, and Alexander Hoffman.
\newblock Accelerating deep learning model inference on arm cpus with ultra-low
  bit quantization and runtime.
\newblock {\em CoRR}, abs/2207.08820, 2022.

\bibitem{bhalgat2020lsq+}
Yash Bhalgat, Jinwon Lee, Markus Nagel, Tijmen Blankevoort, and Nojun Kwak.
\newblock {LSQ+:} improving low-bit quantization through learnable offsets and
  better initialization.
\newblock {\em CoRR}, abs/2004.09576, 2020.

\bibitem{cai2017hwgq}
Zhaowei Cai, Xiaodong He, Jian Sun, and Nuno Vasconcelos.
\newblock Deep learning with low precision by half-wave gaussian quantization.
\newblock {\em CoRR}, abs/1702.00953, 2017.

\bibitem{chen2018tvm}
Tianqi Chen, Thierry Moreau, Ziheng Jiang, Lianmin Zheng, Eddie Yan, Haichen
  Shen, Meghan Cowan, Leyuan Wang, Yuwei Hu, Luis Ceze, et~al.
\newblock $\{$TVM$\}$: An automated end-to-end optimizing compiler for deep
  learning.
\newblock In {\em 13th $\{$USENIX$\}$ Symposium on Operating Systems Design and
  Implementation ($\{$OSDI$\}$ 18)}, pages 578--594, 2018.

\bibitem{choi2018sawb}
Jungwook Choi, Pierce~I{-}Jen Chuang, Zhuo Wang, Swagath Venkataramani,
  Vijayalakshmi Srinivasan, and Kailash Gopalakrishnan.
\newblock Bridging the accuracy gap for 2-bit quantized neural networks
  {(QNN)}.
\newblock {\em CoRR}, abs/1807.06964, 2018.

\bibitem{choi2019pactsawb}
Jungwook Choi, Swagath Venkataramani, Vijayalakshmi~(Viji) Srinivasan, Kailash
  Gopalakrishnan, Zhuo Wang, and Pierce Chuang.
\newblock Accurate and efficient 2-bit quantized neural networks.
\newblock In A. Talwalkar, V. Smith, and M. Zaharia, editors, {\em Proceedings
  of Machine Learning and Systems}, volume~1, pages 348--359, 2019.

\bibitem{choi2018pact}
Jungwook Choi, Zhuo Wang, Swagath Venkataramani, Pierce~I{-}Jen Chuang,
  Vijayalakshmi Srinivasan, and Kailash Gopalakrishnan.
\newblock {PACT:} parameterized clipping activation for quantized neural
  networks.
\newblock {\em CoRR}, abs/1805.06085, 2018.

\bibitem{Cowan2020}
Meghan Cowan, Thierry Moreau, Tianqi Chen, James Bornholt, and Luis Ceze.
\newblock Automatic generation of high-performance quantized machine learning
  kernels.
\newblock In {\em Proceedings of the 18th ACM/IEEE International Symposium on
  Code Generation and Optimization}, CGO 2020, page 305–316, New York, NY,
  USA, 2020. Association for Computing Machinery.

\bibitem{dukhan2018qnnpack}
Marat Dukhan, Yiming Wu, and Hao Lu.
\newblock {QNNPACK: open source library for optimized mobile deep learning},
  2018.

\bibitem{esser2019lsq}
Steven~K. Esser, Jeffrey~L. McKinstry, Deepika Bablani, Rathinakumar Appuswamy,
  and Dharmendra~S. Modha.
\newblock Learned step size quantization.
\newblock {\em CoRR}, abs/1902.08153, 2019.

\bibitem{Han2020}
Qingchang Han, Yongmin Hu, Fengwei Yu, Hailong Yang, Bing Liu, Peng Hu, Ruihao
  Gong, Yanfei Wang, Rui Wang, Zhongzhi Luan, and Depei Qian.
\newblock {Extremely Low-bit Convolution Optimization for Quantized Neural
  Network on Modern Computer Architectures}.
\newblock {\em ACM International Conference Proceeding Series}, 2020.

\bibitem{hu2018bitflow}
Yuwei Hu, Jidong Zhai, Dinghua Li, Yifan Gong, Yuhao Zhu, Wei Liu, Lei Su, and
  Jiangming Jin.
\newblock Bitflow: Exploiting vector parallelism for binary neural networks on
  cpu.
\newblock In {\em 2018 IEEE International Parallel and Distributed Processing
  Symposium (IPDPS)}, pages 244--253. IEEE, 2018.

\bibitem{jacob2017gemmlowp}
Benoit Jacob et~al.
\newblock gemmlowp: a small self-contained low-precision gemm library.(2017),
  2017.

\bibitem{Lai2018}
Liangzhen Lai, Naveen Suda, and Vikas Chandra.
\newblock {CMSIS-NN:} efficient neural network kernels for arm cortex-m cpus.
\newblock {\em CoRR}, abs/1801.06601, 2018.

\bibitem{li2021mqbench}
Yuhang Li, Mingzhu Shen, Jian Ma, Yan Ren, Mingxin Zhao, Qi Zhang, Ruihao Gong,
  Fengwei Yu, and Junjie Yan.
\newblock Mqbench: Towards reproducible and deployable model quantization
  benchmark.
\newblock {\em CoRR}, abs/2111.03759, 2021.

\bibitem{paszke2019pytorch}
Adam Paszke, Sam Gross, Francisco Massa, Adam Lerer, James Bradbury, Gregory
  Chanan, Trevor Killeen, Zeming Lin, Natalia Gimelshein, Luca Antiga, Alban
  Desmaison, Andreas K{\"{o}}pf, Edward~Z. Yang, Zach DeVito, Martin Raison,
  Alykhan Tejani, Sasank Chilamkurthy, Benoit Steiner, Lu Fang, Junjie Bai, and
  Soumith Chintala.
\newblock Pytorch: An imperative style, high-performance deep learning library.
\newblock {\em CoRR}, abs/1912.01703, 2019.

\bibitem{sankaran2021neutrino}
Anush Sankaran, Olivier Mastropietro, Ehsan Saboori, Yasser Idris, Davis
  Sawyer, MohammadHossein AskariHemmat, and Ghouthi~Boukli Hacene.
\newblock Deeplite neutrino: An end-to-end framework for constrained deep
  learning model optimization.
\newblock {\em CoRR}, abs/2101.04073, 2021.

\bibitem{ncnn}
Tencent.
\newblock {NCNN, A high-performance neural network inference framework
  optimized for the mobile platform}, 2020.

\bibitem{Tulloch2017}
Andrew Tulloch and Yangqing Jia.
\newblock High performance ultra-low-precision convolutions on mobile devices.
\newblock {\em CoRR}, abs/1712.02427, 2017.

\bibitem{ulppack}
Jaeyeon Won, Jeyeon Si, Sam Son, Tae~Jun Ham, and Jae~W. Lee.
\newblock Ulppack: Fast sub-8-bit matrix multiply on commodity simd hardware.
\newblock In D. Marculescu, Y. Chi, and C. Wu, editors, {\em Proceedings of
  Machine Learning and Systems}, volume~4, pages 52--63, 2022.

\bibitem{yamamoto2021lcq}
Kohei Yamamoto.
\newblock Learnable companding quantization for accurate low-bit neural
  networks.
\newblock {\em CoRR}, abs/2103.07156, 2021.

\bibitem{yao2020hawq}
Zhewei Yao, Zhen Dong, Zhangcheng Zheng, Amir Gholami, Jiali Yu, Eric Tan,
  Leyuan Wang, Qijing Huang, Yida Wang, Michael~W. Mahoney, and Kurt Keutzer.
\newblock {HAWQV3:} dyadic neural network quantization.
\newblock {\em CoRR}, abs/2011.10680, 2020.

\bibitem{zhou2016dorefa}
Shuchang Zhou, Zekun Ni, Xinyu Zhou, He Wen, Yuxin Wu, and Yuheng Zou.
\newblock Dorefa-net: Training low bitwidth convolutional neural networks with
  low bitwidth gradients.
\newblock {\em CoRR}, abs/1606.06160, 2016.

\end{thebibliography}
}

\end{document}